\documentclass{article}
\usepackage{ijcai23}

\usepackage{times}
\usepackage{soul}
\usepackage{url}
\usepackage[hidelinks]{hyperref}
\usepackage[utf8]{inputenc}
\usepackage[small]{caption}
\usepackage{graphicx}
\usepackage{amsmath}
\usepackage{amsthm}
\usepackage{booktabs}
\usepackage{algorithm}
\usepackage{algorithmic}
\usepackage[switch]{lineno}
\usepackage{xcolor}
\usepackage{multirow}
\usepackage{amsfonts}

\makeatletter

\definecolor{c1}{rgb}{0.8235, 0.9098, 0.8901}
\definecolor{c2}{rgb}{1.0000, 0.7960, 0.6039}
\definecolor{c3}{rgb}{0.8470, 0.6901, 0.5490}
\definecolor{c4}{rgb}{0.6196, 0.4117, 0.2274}

\newcommand{\ostar}{\mathbin{\mathpalette\make@circled\star}}
\newcommand{\make@circled}[2]{%
  \ooalign{$\m@th#1\smallbigcirc{#1}$\cr\hidewidth$\m@th#1#2$\hidewidth\cr}%
}
\newcommand{\smallbigcirc}[1]{%
  \vcenter{\hbox{\scalebox{0.77778}{$\m@th#1\bigcirc$}}}%
}
\makeatother

\title{SGHormer: An Energy-Saving Graph Transformer Driven by Spikes}

\author{
Huizhe Zhang$^1$
\and
Jintang Li$^1$\and
Liang Chen$^{1}$ \thanks{Corresponding author.}\And
Zibin Zheng$^1$
\affiliations
$^1$Sun Yat-sen University\\
\emails
\{zhanghzh33, lijt55\}@mail2.sysu.edu.cn,
\{chenliang6, zhzibin\}@mail.sysu.edu
}

\begin{document}

\maketitle

\begin{abstract}

Graph Transformers (GTs) with powerful representation learning ability make a huge success in wide range of graph tasks. However, the costs behind outstanding performances of GTs are higher energy consumption and computational overhead. The complex structure and quadratic complexity during attention calculation in vanilla transformer seriously hinder its scalability on the large-scale graph data. Though existing methods have made strides in simplifying combinations among blocks or attention-learning paradigm to improve GTs' efficiency, a series of energy-saving solutions originated from biologically plausible structures are rarely taken into consideration when constructing GT framework. To this end, we propose a new spiking-based graph transformer (SGHormer). It turns full-precision embeddings into sparse and binarized spikes to reduce memory and computational costs. The spiking graph self-attention and spiking rectify blocks in SGHormer explicitly capture global structure information and recover the expressive power of spiking embeddings, respectively. In experiments, SGHormer achieves comparable performances to other full-precision GTs with extremely low computational energy consumption. The results show that SGHomer makes a remarkable progress in the field of low-energy GTs. Code is available at \url{https://github.com/Zhhuizhe/SGHormer}.
\end{abstract}

\section{Introduction}
Graph neural networks (GNNs) as a flourishing representation learning methods on graph data have been developed and applied on diverse tasks ~\cite{ying2018recommender}~\cite{zhu2023devil}. Most GNNs based on message passing paradigm can effectively generate representations of nodes by exchange the local structure information among nodes ~\cite{hamilton2018inductive}. Despite message passing neural networks (MPNNs) have strong capabilities in capturing graph inductive biases, there are still some of inherent drawbacks are uncovered and formalized such as over-squashing, long-range dependencies and expressive power limitations ~\cite{dai2018longrange}~\cite{xu2019powerful}. Motivated by sequence data modeling and contextual understanding of Transformers, some studies  strive to construct Graph Transformers (GTs) to incorporate global nodes' semantic with local structural information. The self-attention mechanism from the vanilla Transformer calculates attention scores among tokens' pair and outputs the fusion of embeddings containing all attention information. This mechanism can be performed on the graph naturally, which tokens are considered as nodes and attention scores are seen as the importance of edges between two nodes. The process of aggregating different embeddings with attention weights also can be considered as a special case of applying the message-passing rule on a fully-connected graph. Some emerging GTs already achieve competitive or even surpassing performances against MPNNs in many graph tasks ~\cite{ying2021graphormer}~\cite{kreuzer2021san}.

\begin{figure}[t]
\centering
  \includegraphics[width=\linewidth]{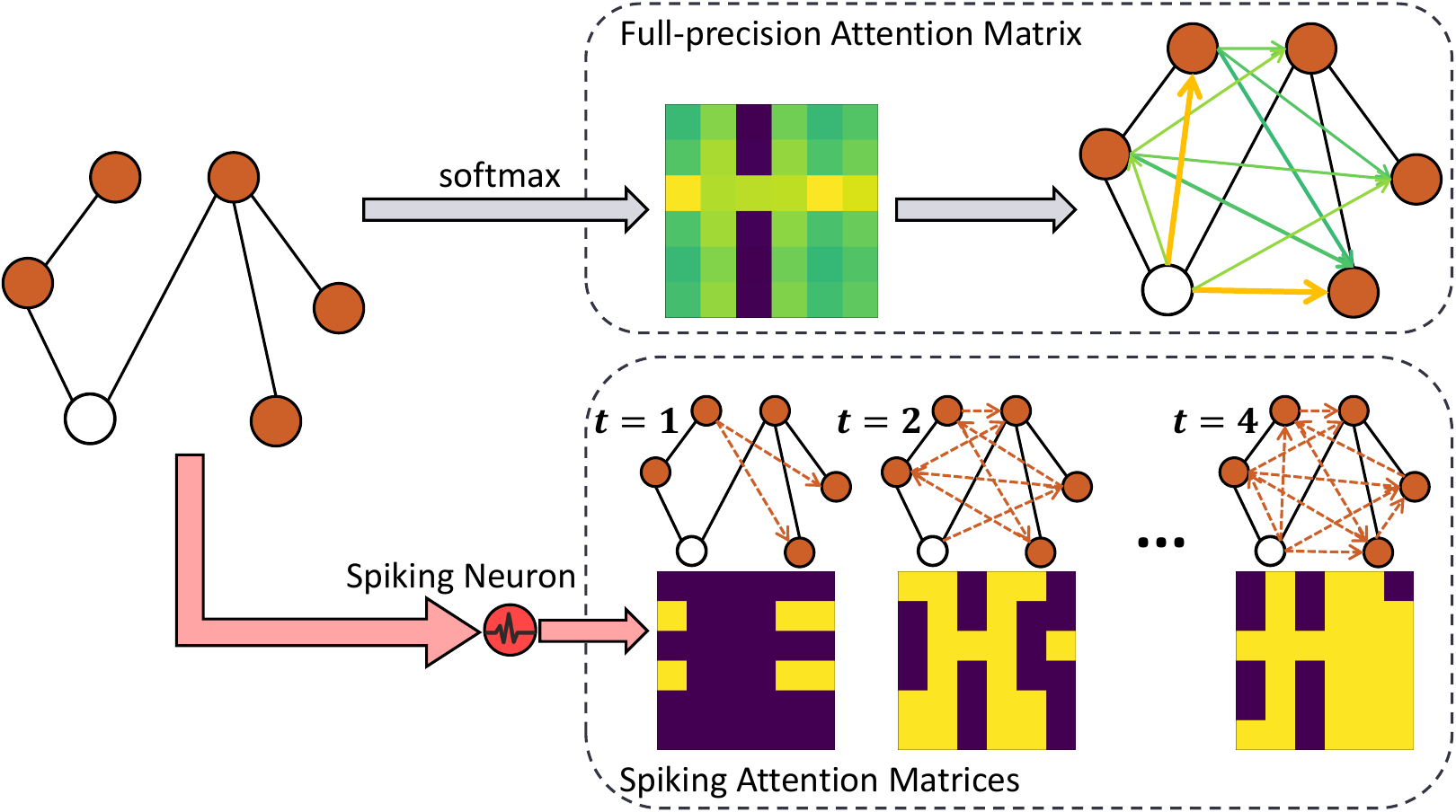}
\caption{Visualization results of spiking attention matrices and the full-precision vanilla self-attention matrix. We construct the experiments on a selected graph from ZINC. For spiking attention matrices cross multiple time steps ($\textbf{bottom}$), parts of spiking outputs have similar attention patterns as the full-precision attention ($\textbf{top}$) calculated by a softmax function.}
\label{fig:1}
\end{figure}

However, integrating Transformers and MPNNs to fully unleash their representational power on the large-scale graph data still faces some challenges. Firstly, the scheme is still ambiguous that injecting explicit connectivity information into Transformers to alleviate its deficiency in lack of strong inductive biases ~\cite{ma2023graph}. Because the self-attention tends to ignore the structural information or relations between node pairs, it make Transformers hard to generate meaningful attention scores from the view of the graph topology information. Besides, for addressing these issues, those GTs with complex attention mechanism always lead to astonishing computational energy consumption. As we mentioned before, the vanilla self-attention mechanism is equivalent to calculate weights of every nodes' pair in a fully-connected graph. The computation and memory costs behind similar operations on the large-scale graph are unacceptable. Some approaches which attempt to improve the efficiency of GTs are still in early stages, and new effective solutions are yet to be explored further.

With development of neuromorphic hardwares, spiking neural networks (SNNs), as biologically plausible structures, have potential to break through the energy consumption bottleneck of Transformers. The brain-inspired architecture have attracted widespread attention as a new-generation efficient neural networks. Different from the neurons in artificial neural networks (ANNs), biological neurons interact with each other with sparse spikes ~\cite{eshraghian2023lesson}. Recently, the huge advancements in training algorithms enable SNNs composed by neurons to borrow sophisticated architectures from ANNs while further improve the energy efficiency of models. Some of them are increasingly becoming the cornerstone in different applications such as image recognition, gesture recognization and robot control~\cite{roy2019towards}~\cite{zhou2023spikingformer}. Furthermore, it is also an effective way to utilize the sparse and binarized spikes output from biological neurons design a more lightweight, energy-efficient GNNs. Some pioneers already inject SNN into graph models and verify advantages of the fusion architecture on energy efficiency ~\cite{xu2021exploiting}~\cite{zhu2022spiking}~\cite{li2023spikenet}. The experimental results show
promising potentiality of generalizing biologically plausible networks on the graph data, SNNs have been underappreciated and under-investigated in GTs. Introducing biological neurons into current Transformer frameworks may likely offer a sustainable, low-energy solution for GTs.

Based on SNNs, the sparse operations designed for spiking neurons bring huge reductions in memory and consumption costs. Simultaneously, as shown in Figure \ref{fig:1}, we observe that replacing the softmax function with a spiking neuron, self-attention blocks may capture similar attention patterns. Due to the regularity of spikes, it is possible to learn attention patterns using spiking attention matrices from few time steps. These characteristics drive us to integrate the spiking neurons with GTs. 
In this study, we construct a spiking driven graph transformer (SGHormer). As far as we know, it is also the first methods which inject the spiking neurons into GTs. Specifically, there are two main components in SGHormer, spiking rectify block (SRB) and spiking graph self-attention (SGSA). SRBs recover and generate the approximated input embeddings, which can effectively alleviate the information loss during spiking. Based on observations, SGSA not only alleviates the problem about dependencies of SNNs on time steps, but also generates the spiking attention matrix in a power-efficient way. The contributions of this paper are summarized as follows:

The following instructions apply to submissions:
\begin{itemize}
\item We create an energy-saving graph transformer framework using biologically inspired spiking neurons.

\item For this new spiking-driven GTs, We design Spiking Graph Attention Head (SGSA) and Spiking Rectify Blocks (SRB), which effectively utilize the inherent filer operation to simplify the self-attention calculation and alleviate the strong dependencies of spiking neurons on time steps.

\item We compare our methods with 5 advanced GTs and 8 GNNs on PYG and OGB datasets. And we also compare the theoretical energy consumption of our models with that of other GTs. The results show that SGHormer can achieve comparable performances against other full-precision GTs with extremely low computational energy consumption.

\end{itemize}

\section{Related Work}
\paragraph{Spiking Neural Networks.} Spiking Neural Networks (SNNs), characterized by low power consumption, event-driven features and biological plausibility, are considered the third generation of neural networks. Motivated by brain's neural circuitry, neurons in a SNN communicate with spikes which can be seen as electrical impulses when membrane potential reach the threshold. Early spiking neurons like Hodgkin-Huxley Model follow a biophysical mechanism that currents caused by action potentials will go through ion channels in the cell membrane ~\cite{gerstner2014neuronal}. Starting from the Hodgkin-Huxley model, the derivation of simplified neuron models such as IF and LIF are proposed for adapting deep learning framework. Though output discrete, single-bit spikes extremely improve the efficiency of neural networks, binarized outputs also raise the non-differentiable problem which makes challenging for directly training SNNs through the backpropagation algorithm. ANN-to-SNN and surrogate gradients are two common ways to relieve the above questions ~\cite{cao2015spiking}~\cite{zhou2022spikformer}. There are lots of studies in the field of computer vision show that surrogate gradients can achieve approximated performances compared with the normal gradient decent.
 
\paragraph{Graph Transformers.} As a transformative framework, Transformer and its variants achieve tremendous success and gradually become new benchmarks in various domains ~\cite{vaswani2017attention}~\cite{dosovitskiy2021image}. Due to self-attention mechanism can naturally be regarded as special case for importance calculation for nodes' pairs on the graph data. Emerged GTs verify this assumption. Some works focus on integrate the original MPNN with new Transformer framework ~\cite{rampasek2022gps}~\cite{ma2023graph}. Other methods choose to further modify the attention calculation for reducing the quadratic complexity into linear complexity ~\cite{wu2022nodeformer}~\cite{wei2023lightgt}. 
In this work, we aim to build a spiking Graph Transformer architecture from the view of neuroscience and explore future Transformer-based neuromorphic chip design. 

\paragraph{Positional/Structural encodings.} Recent studies indicate that manually constructing positional and structural encodings (PE/SE) contributes to make standard MPNN more expressive than 1-Weisfeiler-Leman test ~\cite{srinivasan2020equivalence}. There are two common encoding strategies, positional and structural encodings. Positional encodings is mainly used to generate embeddings that contain information about the location of nodes in the graph. There are many kinds of PE have been developed, for example, Laplacian PE or Weisfeiler-Lehman-based PE ~\cite{li2020de}~\cite{dwivedi2021generalization}. The examples of structural encodings include degree of a node, random-walk SE and so on ~\cite{ying2021graphormer}~\cite{dwivedi2022graph}. A common way to inject the supplementary information is adding or concatenating the positional or structural representations of the graph to with node features before the main Transformer model. In this work, we directly incorporate Laplacian PE and random-walk SE to generate auxiliary graph topological information.

\begin{figure*}
    \centering
    \includegraphics[width=0.9\textwidth]{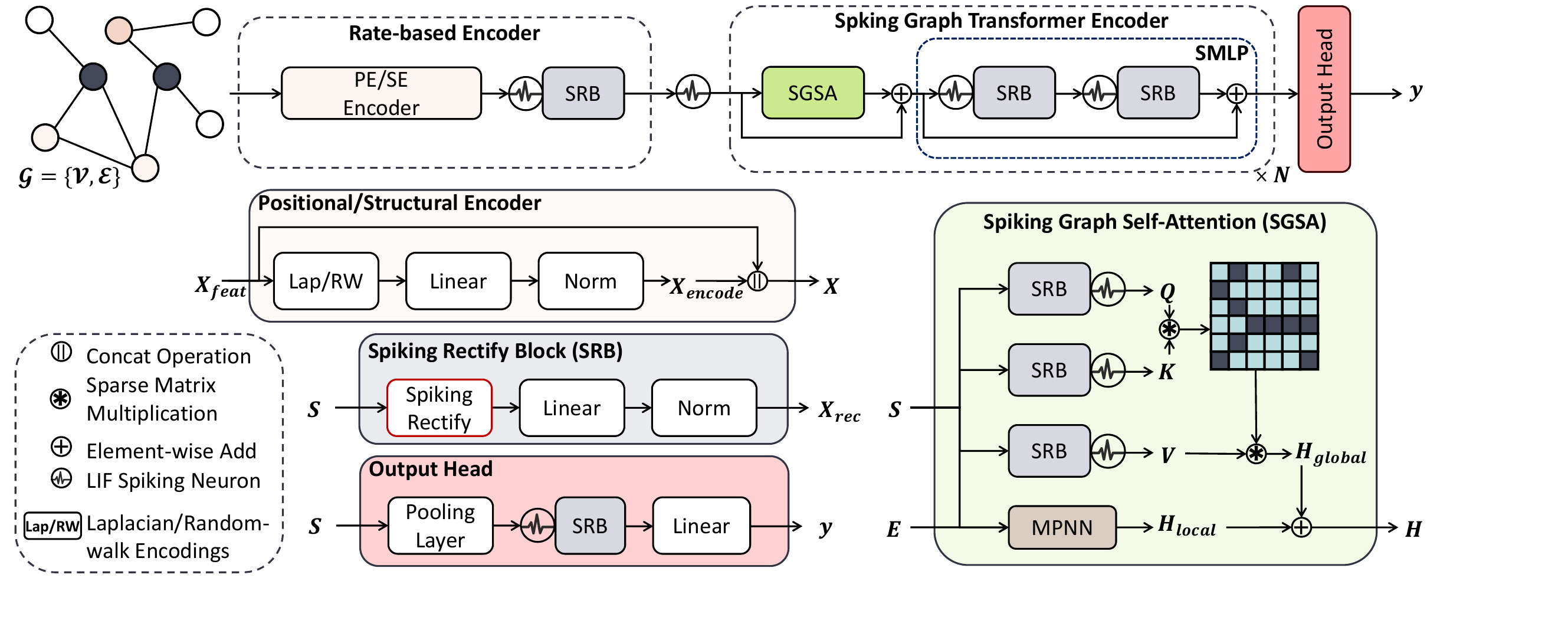}
    \caption{The framework of SGHormer}
    \label{fig:2}
\end{figure*}

\section{Preliminaries}
\paragraph{Spiking neural networks.}  Though the electrophysiological measurements can be calculated accurately by those complex conductance-based neurons, the complexity also limits widespread deployment in deep neural networks. Currently, most of SNNs consist of the simpler computational units, IF, LIF and PLIF ~\cite{gerstner2014neuronal}. Specifically, a spiking neuron receives the weighted sum of input current and accumulates membrane potential. Subsequently, the neuron  compared its membrane potential with a threshold to determine whether to generate the spikes. The membrane potential activity can be formulated as follows:
\begin{equation}
\label{eq:1}
    \tau \frac{dV^t}{dt}=-(V^t) + RI^t,
\end{equation}
where $V^t$ is the membrane potential, $\tau$ denotes a time constant of membrane and $R$ denotes as the membrane resistance. For facilitating the deployment of biological neurons in deep learning, the membrane potential reset and spiking should be retained while relaxing the physically viable assumptions. To this end, Eq. \ref{eq:1} can be converted into an iterative expression as follows:
\begin{equation}
    \label{eq:2}
    V^t=V^{t-1} + \beta(WX^t - (V^{t-1} - V_{reset})),
\end{equation}
\begin{equation}
    \label{eq:3}
    V^{t} = V^t(1 - S^t) + V_{reset} S^t,
\end{equation}
\begin{equation}   
    \label{eq:4}
    S^t = \left\{
    \begin{array}{ll}
         1,& V^t \geq V_{th} \\
         0,& otherwise
    \end{array},
    \right.
\end{equation}
where $\beta$ can be considered as simplified a decay constant. We utilize $snn(\cdot)$ to denote a LIF neuron hereafter. And we use the surrogate gradient to approximate gradients of parameters as follows:
\begin{equation}   
    \label{eq:5}
    \frac{\partial \mathcal{L}}{\partial S}
    \frac{\partial S}{\partial V}
    \frac{\partial V}{\partial I}
    \frac{\partial I}{\partial W}\approx
    \frac{\partial \mathcal{L}}{\partial S}
    \frac{\partial \tilde{S}}{\partial V}
    \frac{\partial V}{\partial I}
    \frac{\partial I}{\partial W},
\end{equation}
where $\tilde{S}$ is function substitution. In this work, we also employ spike gradients, which enables the model to reach the convergence quickly even with low time steps.
\paragraph{Self-attention.} One of the most prominent components of Transformer is multi-head self-attention mechanism. For $\mathcal{G}=(\mathcal{V},\mathcal{E})$, and $X\in\mathbb{R}^{N \times d}$ denotes the nodes' features. let $X\in\mathbb{R}^{N \times d}$ be the input to a vanilla self-attention layer, where $N$ is the number of nodes in a graph and $d$ is dimensions of hidden embeddings. Query, key and value matrices are calculated by corresponding learnable projection matrices respectively, as defined below:
\begin{equation}
\label{eq:6}
    Q=XW_q,K=XW_k,V=XW_v,
\end{equation}
where $W_q,W_k,W_v\in\mathbb{R}^{d \times d'}$. For $Q$ and $K$, the layer calculates dot products of a query with all keys. Results are divided by $\sqrt{d'}$ and fed into a softmax function to calculate attention scores of each value. At last, for multi-head self-attention, the concatenation of output embeddings from $M$ different heads will be integrated and applied a linear transformation:
\begin{equation}
\label{eq:7}
    H_m=softmax(\frac{Q_mK_m^T}{\sqrt{d'}})V_m,
\end{equation}
\begin{equation}
\label{eq:8}
    H=Linear(H_1 \parallel H_2 \parallel ... \parallel H_M),
\end{equation}
where $\parallel$ denotes concat operations.
\section{Methodology}
In this section, we detail a new spike-based transformer framework to change attention calculation into a graph reconstruction by spiking neurons. The framework of SGHormer is shown in Figure \ref{fig:2}. As depicted in figure, SGHormer changes position encodings and corresponding attributes of nodes from continuous and full-precision values into rate coded spikes by rate-based encoder. Then spikes are send into the spiking graph self-attention block. SGSA binarizes global attention scores which turn the attention calculate task into a graph reconstruction task dominated by spiking neurons. Simultaneously, SGSA create the local embeddings of nodes using explicit connectivity information. At last, the sparse spiking embeddings are fed into a output head to generate the corresponding predictive results for downstream classification or regression tasks.

\subsection{Rate-coded nodes' feature}
As a rate-based SNN, we follow the same hypothesis that the spiking rate is proportional to the importance of patterns in nodes' features ~\cite{zhu2022spiking}. The higher intensity of features is equal to a higher spike count or spiking rate in the same time interval $T$. Let $X_{feat}$ and $X_{encode}$ be input node features and positional/structural encodings, respectively. The concatenation of above embedding $X$ are fed into a rate encoder to generate rate-coded multi-temporal spikes $S=\{S^1, S^2,...,S^T\}$. There are two common options to turn inputs into rate coding spikes. For a probability-based encoder, it consider the rate encoding as a Bernoulli trial. Another approach repeatedly pass the embeddings of nodes into a shared spiking neuron $T$ times. We choose the latter as rate-based encoder in SGHormer. The process can be defined by the following formulations:
\begin{equation}
\label{eq:9}
   S_{i}^t=snn(X_i)=snn(Linear(X_{i,feat} \parallel X_{i, encode})),
\end{equation}
where $X_i$ is the integrated embedding of node $i$, $S_{i}^t$ denotes output spikes at the $t$-th time step. Calculating the spiking rate along $T$ time steps can yield approximate estimate of raw embeddings before entering spiking neurons.

\subsection{Spiking rectify block}
As we mentioned before, all layers in SGHormer communicate with each other only using sparse and binarized spikes. It means that there are lots of operations involving the conversion from full-precision values to binary spikes. While some attempts demonstrate that constructing a deeper networks can also improve performances of SNNs ~\cite{fang2021sew}~\cite{zheng2021deepersnn}~\cite{hu2023sresnet}, the negative impact of information loss becomes prominent on a more complex model and those advantages brought by deeper network framework will evaporate. Hence, we design a extra spiking rectify block (SRB) to recover input raw embeddings with real values from output spikes. We assume a raw embedding $x_i$ follow the normal distribution. Based on the output spiking vector $s_i$, the mean $\hat{\mu}$ and variance $\hat{\sigma}$ of spikes across multiple time steps can be calculated. Subsequently, SRB utilizes the approximate results and rectifies output spikes filling values. The specific process can be formulated as follows:
\begin{equation}
\label{eq:10}    
   \hat{\mu}_i = \frac{1}{T}\sum_{t=1}^Ts_{i}^t,
   \hat{\sigma}_i = \frac{1}{T}\sum_{t=1}^T(s_{i}^t-\hat{\mu}_i)^2,
\end{equation}
\begin{equation}
\label{eq:11}
   \hat{X}^t=S^t-W\odot U^t,U^\sim N(\hat{\mu},1-\hat{\sigma}),
\end{equation}
\begin{equation}
\label{eq:12}
   X_{rec}^t = rec(S^t)=BN(Linear(\hat{X}^t),
\end{equation}
where $W^t$ is a learnable weight matrix, $\odot$ is the Hadamard product. As shown in Figure \ref{fig:2}, SRB utilizes the correlation between output spikes and input embeddings to reconstruct nodes' embedding rather than apply linear transformation and normalization directly. 

\begin{table*}
    \centering
    \begin{tabular}{lccccc}
        \toprule
        \multirow{2}*{Model}  & ZINC & MNIST & CIFAR10 & PATTERN & CLUSTER \\
        \cline{2-6}
        & MAE$\downarrow$ & Accuracy$\uparrow$ & Accuracy$\uparrow$ & Accuracy$\uparrow$ & Accuracy$\uparrow$ \\
        \midrule
        GCN & 0.367$\pm$0.011  & 90.705$\pm$0.218  & 55.710$\pm$0.381  & 71.892$\pm$0.334  & 68.498$\pm$0.976  \\
        GIN & 0.526$\pm$0.051  & 96.485$\pm$0.252  & 55.255$\pm$1.527  & 85.387$\pm$0.136  & 64.716$\pm$1.553  \\
        GAT & 0.384$\pm$0.007  & 95.535$\pm$0.205  & 64.223$\pm$0.455  & 78.271$\pm$0.186  & 70.587$\pm$0.447  \\
        GatedGCN & 0.282$\pm$0.015  & 97.340$\pm$0.143  & 67.312$\pm$0.311  & 85.568$\pm$0.088  & 73.840$\pm$0.326  \\
        GatedGCN-LSPE & 0.090$\pm$0.001  & -             & -             & -             & -  \\
        PNA & 0.188$\pm$0.004  & 97.940$\pm$0.12   & 70.350$\pm$0.630  & -             & -  \\
        DGN & 0.168$\pm$0.003  & -             & \textbf{\textcolor{red}{72.838$\pm$0.417}}  & 86.680$\pm$0.034  & -  \\
        GSN & 0.101$\pm$0.010  & -             & -             & -             & -  \\
        \midrule
        SAN & 0.139$\pm$0.006  & -             & -             & 86.581$\pm$0.037  & 76.691$\pm$0.650  \\
        Graphormer & 0.122$\pm$0.006  & -             & -             & -             & -  \\
        K-Subgraph SAT & 0.094$\pm$0.008 & -             & -             & \textbf{\textcolor{red}{86.848$\pm$0.037}}  & 77.856$\pm$0.104  \\
        EGT & 0.108$\pm$0.009  & 98.051$\pm$0.126  & 68.702$\pm$0.409  & 86.821$\pm$0.020  & \textbf{\textcolor{red}{79.232$\pm$0.348}}  \\
        GPS & \textbf{\textcolor{red}{0.070$\pm$0.004}}  & \textbf{\textcolor{red}{98.173$\pm$0.087}}  & 72.298$\pm$0.356  & 86.685$\pm$0.059  & 78.016$\pm$0.180  \\
        \midrule
        ours & \colorbox{c2}{0.117$\pm$0.032} & \colorbox{c3}{96.850$\pm$0.247} & \colorbox{c4}{67.740$\pm$0.158} & \colorbox{c1}{86.527$\pm$0.620} & \colorbox{c4}{71.279$\pm$0.205}  \\ 
        \bottomrule
    \end{tabular}
    \caption{Test performances on five Benchmarking-GNNs datasets. The best result are highlighted in \textbf{\textcolor{red}{red}}. Color blocks represent the difference between SGHormer and other full-precision methods, with darker colors indicating larger performance gaps.}
    \label{tab:1}
\end{table*}

\begin{table}
    \centering
    \begin{tabular}{lcc}
        \toprule
        \multirow{2}*{Model}  & ogbg-molhiv & ogbg-molpcba \\
        \cline{2-3}
        & AUC$\uparrow$ & AP$\uparrow$ \\
        \midrule
        GCN+virtual node & 75.90$\pm$1.1 & 24.24$\pm$0.3 \\
        GIN+virtual node & 77.07$\pm$1.4 & 27.03$\pm$0.2 \\
        GatedGCN-LSPE    & -             & 26.70$\pm$0.2 \\
        PNA        & 79.05$\pm$1.3 & 28.38$\pm$0.3 \\
        DeeperGCN  & 78.58$\pm$1.1 & 27.81$\pm$0.3 \\
        DGN        & 79.70$\pm$0.9 & 28.85$\pm$0.3 \\
        GSN (directional) & \textbf{\textcolor{red}{80.39$\pm$0.9}} & -             \\
        \midrule
        SAN        & 77.85$\pm$0.2 & 27.65$\pm$0.4 \\
        GraphTrans (GCN-Virtual) & -             & 27.61$\pm$0.2 \\
        K-Subtree SAT        & -             & -             \\
        GPS        & 78.80$\pm$1.0 & \textbf{\textcolor{red}{29.07$\pm$0.2}} \\
        \midrule
        SGHormer   & \colorbox{c2}{77.47$\pm$0.4} & \colorbox{c3}{27.43$\pm$0.1} \\ 
        \bottomrule
    \end{tabular}
    \caption{Test performances on two OGB datasets.}
    \label{tab:2}
\end{table}
\subsection{Spiking graph attention head} 
Different from the vanilla transformer, GTs not only need capture the local connectivity information utilizing explicit edges in the graph, but also calculate global attention scores to infer relation between each nodes' pair ~\cite{ying2021graphormer}~\cite{rampasek2022gps}. Benefit from sparse and binarized spikes, spiking graph self-attention (SGSA) contain MPNN and self-attention blocks to undertake above two objectives with low computation and memory overhead. Specifically, we fed the query, key and value matrices into different spiking neurons. Output matrices after spiking neurons is sparse and binary, which only contain 0/1 elements. Therefore, the linear operation on these spiking matrices is only addition. The matrix multiplication in attention computation also can be transformed into a sparse form to further reduce memory consumption. On the basis of above sparse operations, SGSA generates node embeddings with local and global structure information. Besides, consistent with the original self-attention architecture, we construct a spiking multilayer perceptron (SMLP) to project outputs into the embedding space. The combination of above blocks can be written as:
\begin{align}
\label{eq:13}
    H_{local}^l={MPNN}(S^l,E),
\end{align}
\begin{align}
\label{eq:14}
    \begin{split}
    H_{global}^l=\tilde{A}^lV^l&=\Theta(Q_{sp}^l, K_{sp}^l)V_{sp}^l\\
    &=(g(Q^l) \ostar g(K^l))g(V^l)
    \end{split},
\end{align}
\begin{equation}
\label{eq:15}
    S^{l+1}=SMLP(H_{local}^l + H_{global}^l),
\end{equation}
where $g(\cdot)=snn(rec(\cdot))$, $\ostar$ denotes the sparse matrix multiplication. It also can be replaced by XNOR and bitcount operations ~\cite{Rastegari2016xnor}. Notably, we just use a plain message passing layer to aggregate neighbors' and edges' feature, which can be defined as:
\begin{equation}
\label{eq:16}
    h_i=MPNN(x_i, e)=x_i+\sum_{j\in \mathcal{N}(i)}(wx_j + e_{ij}),
\end{equation}
where $\mathcal{N}(\cdot)$ denotes the immediate neighborsof node, $e_{ij}$ is the edge feature between nodes $i$ and $j$. As shown in Eq. \ref{eq:14}, The global attention scores are calculated by query and key spiking matrices. From this perspective, the task about inferring latent relations on a fully-connected graph can be considered as the graph reconstruction controlled by spiking neurons. Because the nonnegativity of spikes, we further remove the softmax function to simplify the computation of self-attention. For any given self-attention layer, a simple way to compute the global embedding $H=\{H^{1},H^{2},...,H^{T}\}$ for $T$ time steps can be written as:
\begin{equation}
\label{eq:17}
H^{1}=\tilde{A}^{1}V^{1},H^{2}=\tilde{A}^{2}V^{2},...,H^{T}=\tilde{A}^{T}V^{T},
\end{equation}
However, we believe that such operations overlook the inherent filter operations of SNNs. And the arrival times of spikes can also reveal the intensity between different elements using the rate-coding method mentioned in the previous section. The earlier a spike arrives, the greater its input value compared to the membrane potential threshold of a spiking neuron. Therefore, we consider an extreme scenario that the most important latent relations emerges in $\tilde{A}^{1,l}$ after filtering by spiking neuron. At this point, the process of generating global embedding can be updated as: 
\begin{equation}
\label{eq:18}
H^{1}=\tilde{A}^{1}V^{1},H^{2}=\tilde{A}^{1}V^{2},...,H^{T}=\tilde{A}^{1}V^{T},
\end{equation}
\begin{equation}
\label{eq:19}
\tilde{A}^{1}=g(Q^{1}) \ostar g(K^{1})),
\end{equation}
This simplified operation significantly reduces the dependence of SNNs on long time steps. Besides, it still ensuring that the global connectivity information can be captured by SGSA. We choose the latter one in our experiments.

\begin{table*}
    \centering
    \begin{tabular}{lcccccccccccc}
        \toprule
        \multirow{3}*{Model}  & \multicolumn{3}{c}{ZINC} & \multicolumn{3}{c}{MNIST} & \multicolumn{3}{c}{PATTERN} & \multicolumn{3}{c}{ogbg-molhiv} \\
        \cmidrule(r){2-4} \cmidrule(r){5-7} \cmidrule(r){8-10} \cmidrule(r){11-13}
        & Param & Mem & Eng & Param & Mem & Eng & Param & Mem & Eng & Param & Mem & Eng \\
        & (MB) & (GB) & (mJ) & (MB) & (GB) & (mJ) & (MB) & (GB) & (mJ) & (MB) & (GB) & (mJ) \\
        \midrule
        SAN        & 0.19 & 0.58 & 22.73 & 0.48 & 15.07 & 200.43 & 0.39 & 14.68 & 907.49 & 0.49 & 6.45 & 40.44 \\
        Graphormer & 0.10 & 0.41 & 6.64 & 0.15 & 2.71 & 51.81 & 0.10 & 0.51 & 212.18 & 0.15 & 2.76 & 11.79 \\
        SAT        & 0.24 & 0.52 & 13.65 & 0.36 & 17.21 & 112.99 & 0.24 & 2.97 & 599.0 & 0.37 & 2.05 & 23.38 \\
        GPS        & 0.16 & 0.40 & 6.88 & 0.32 & 15.44 & 52.55 & 0.21 & 0.73 & 214.63 & 0.32 & 2.07 & 12.17 \\
        \midrule
        Average    & 0.17 & 0.48 & 12.47 & 0.33 & 12.61 & 104.44 & 0.24 & 4.72 & 483.33 & 0.33 & 3.33 & 21.94 \\
        SGHormer   & \colorbox{c1}{0.09} & 0.56 & \colorbox{c3}{0.18} & \colorbox{c1}{0.13} & \colorbox{c1}{6.46} & \colorbox{c4}{0.58} & \colorbox{c1}{0.08} & \colorbox{c1}{1.00} & \colorbox{c4}{1.64} & \colorbox{c1}{0.14} & 4.30 & \colorbox{c3}{0.30} \\ 
        \bottomrule
    \end{tabular}
    \caption{The parameter size (MB), memory usage (GB) and theoretical energy consumption (mJ) of various GTs. Color blocks represent the improvement in efficiency compared SGHormer with other full-precision methods, with darker colors indicating larger disparities.}
    \label{tab:3}
\end{table*}

\section{Experiments}
\subsection{Graph classification}
In this section, we compare SGHormer against various message passing neural networks and graph transformers 
on Benchmarking-GNNs~\cite{dwivedi2023benchmark} and OGB~\cite{hu2021open}. For comprehensively validating the effectiveness of SGHormer, selected datasets covering various graph-related tasks such as graph regression, graph classification and node classification. All experiments are conducted on the standard splits of the evaluated datasets. We perform our model on each dataset 5 times with different random seeds to report the mean and standard deviation. All above experiments are conducted on a single NVIDIA RTX 3090 GPU if not explicitly stated otherwise.

\paragraph{Datasets.} We select five datasets including ZINC, MNIST, CIFAR10, PATTERN and CLUSTER from Benchmarking-GNNs to evaluate our method. For the open graph benchmark (OGB), we select two molecular property prediction datasets with different scales, namely ogbg-molhiv and ogbg-molpcba.

\paragraph{Baselines.} As of now, the spiking-related graph transformers or any GNNs used for graph-level tasks has not been located. Therefore, all selected baselines are in full-precision form. One of the main categories among these models are message passing neural networks: GCN~\cite{kipf2017semisupervised}, GIN~\cite{xu2019powerful}, GAT~\cite{velivckovic2017graph}, GatedGCN~\cite{bresson2018residual}, GatedGCN-LSPE~\cite{dwivedi2022graph}, PNA~\cite{corso2020pna}, DGN~\cite{beaini2021dgn}, GSN~\cite{bouritsas2023gsn}. The others are some advanced graph transformers employed in graph-level tasks widely: SAN~\cite{kreuzer2021san}, Graphormer~\cite{ying2021graphormer}, SAT~\cite{chen2022sat}, EGT~\cite{hussain2022egt}, GPS~\cite{rampasek2022gps}.

\paragraph{Overall performance.} The comparative results are demonstrated in Table \ref{tab:1}. It is evident that, despite some gaps compared to state-of-the-art methods based on binary ground truth distances, SGHormer has achieved comparable predictive performance on most datasets through carefully designed network structures. In certain datasets like ZINC, it even outperforms the predictions of full-precision ground truth methods. For message passing neural networks, SGHormer outperforms GCN, GAT and GIN on every datasets. Since the current model only employs a plain MPNN and a simplifying global attention mechanism, its performance remains subpar on datasets with plain node's feature (eg. one-hot encoding) like CLUSTER. Improving performance on such datasets is one of the future directions of our work.

\subsection{Theoretical Energy Consumption}
To examine the energy efficiency of SGHormer, we measure SGHormer and other GTs from three different metrics, model size, memory usage and theoretical energy consumption. However, directly applying the model on neuromorphic chip is rarely explored ~\cite{zhu2022spiking}. To investigate the energy consumption of SGHormer, we derived the theoretical energy consumption from previous works ~\cite{zhou2022spikformer}. Since GTs are still in the early stages of development, there is no standard model size set for each task. For the sake of fairness in comparison, we set the same fixed hyperparameters including the number of layers, the number of heads, the dimension of hidden embeddings for each model while calculating the energy consumption. Besides, different GTs employing various encoding strategies, part of methods preprocess structural information and embed it into node features, while others tend to build a learnable encoder block into the network. We only calculate the energy consumption of transformer encoder in inference step by counting floating point operations (FLOPs) and synaptic operations (SOPs). And the theoretical energy consumption of SGHormer can be formulated as follows:
\begin{equation}
\label{eq:20}
E=E_{coding} + \sum_{l=1}^L E_{trans}
\end{equation}
\begin{align}
\label{eq:21}
    \begin{split}
        E&=\alpha_{f}{FLOP}_{coding} \\
        &+\alpha_{s}\sum_{t=1}^T\sum_{l=1}^L({SOP}_{srb}^{t,l} + {SOP}_{mpnn}^{t,l} + {SOP}_{attn}^{t,l})
    \end{split}
\end{align}
\begin{equation}
\label{eq:23}
    {SOP}^{t, l} = r^{t, l} \times {FLOP}^{t,l}
\end{equation}
where $\alpha_f$ and $\alpha_s$, as scale factors for floating point and synaptic operations, which are set to 4.5 and 0.9, respectively. $r^{t,l}$ is fire rate of block in the $l$-th layer at the $t$-th time step. The theoretical energy consumption results are shown in Table \ref{tab:3}. For all datasets, our method are obviously outperform than the other GTs in size and energy consumption. The average energy consumption is 153x lower compared to other models. This advantage is more pronounced on MNIST and PATTERN which contain more nodes and edges. Besides, the size of our model is smaller which contributes to extend SGHormer to edge devices. Currently, there is still a lack of relevant operations and support for sparse and binarized spikes, making it challenging to demonstrate the model's advantages in terms of memory usage. We believe that further designing customized operators for spiking neural network (SNN) on GPUs will accelerate the development of SNNs.
\begin{table}
    \centering
    \begin{tabular}{lccc}
        \toprule
        Model  & ZINC($\downarrow$) & PATTERN($\uparrow$) & ogbg-molhiv($\uparrow$) \\
        \midrule
        ours   & 0.117 & \textbf{86.527} & 77.473 \\
        \midrule
        - SRB  & 0.129 & 85.803 & 77.241 \\
        + SATT   & 0.114 & 86.518 & 76.004 \\
        \midrule
        + IF   & 0.126 & 86.179 & 76.456 \\
        + PLIF & \textbf{0.109} & 86.267 & \textbf{77.478} \\
        \bottomrule
    \end{tabular}
    \caption{Ablation studies on SGHormer. $-x$ means removing the component $x$ from SGHormer. And $+x$ means replacing the original component in SGHormer with $x$.}
    \label{tab:4}
\end{table}
\subsection{Component analysis}
To elaborately discuss the effectiveness of different components, we construct a series of ablation studies on the SGHormer. Specifically, the experiments primarily assess the impacts of three components including: spiking rectify block, spiking graph self-attention and spiking neuron. There are four comparative methods implemented by removing or replacing one of components. The results are demonstrated in the Table \ref{tab:4}
\paragraph{Spiking rectify block.} As shown in table \ref{tab:4}, removing the spiking rectify block significantly impair predictive performances of SGHormer on three datasets. For a deep spiking neural network, the inputs will become sparser after each layer without any intermediate processing or extra supplementary information. In addition, it bring tons of quantization error that encoding full-precision raw data into binarized spikes. Some existing SNNs are sill strongly dependent on the imprecise spiking representations and directly pass spiking outputs to subsequent processes. We suggest that the limitation make SNNs hard to develop similar the network structure with a vast number of parameters like ANNs. SRBs consider received spikes as a biased data, the blocks attempt to learn a estimator for roughly recovering raw embedding. 
\paragraph{Spiking graph self-attention.} What have been discussed in the previous section is that output spikes of layers follows some certain regulation along $T$ time steps. For those raw inputs that far exceed the threshold of membrane potential, they often emit early and with a higher firing rate under the straightforward rate-based encoder. Because of the regularity of output spikes, we suggest that just using few spiking attention matrices from certain time steps can approximatively capture the essential graph structure information. This assumption can be verified in Table \ref{tab:4}. The self-attention through time described in Eq. \ref{eq:17} (hereafter, SATT) can't provide more valuable relation information among nodes compare with the component which just uses the spiking attention matrix at $1$-st the time step.
\paragraph{Spiking neuron.} Except for the two core components in the SGHormer, we also explore the influences of spiking neurons. Table \ref{tab:4} shows that the predictive performances of PLIF with learnable membrane time constants surpass surpass that of LIF and IF. Due to the spiking attention matrix are control the by corresponding spiking neurons. Currently, There is still no 
a implementation of a common neuron that can be generalized to the different graph data. We believe that developing and designing specialized spiking neurons for GTs or graph-related tasks may further improve the performance of SGHormer.

\begin{figure}
    \centering
    \includegraphics[width=0.48\textwidth]{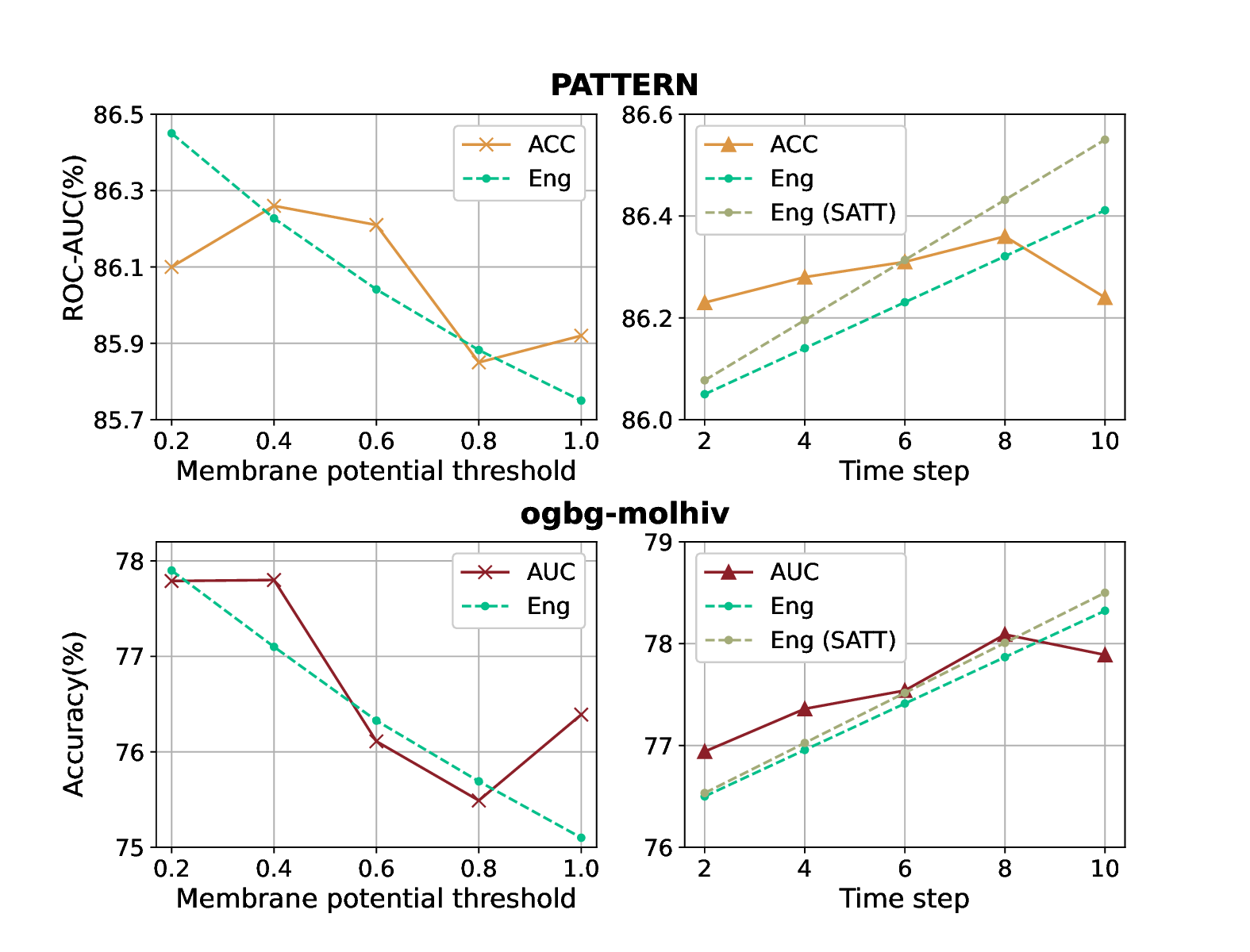}
    \caption{Performances of SGHormer with different membrane potential threshold and time step on PATTERN and ogbg-molhiv datasets.}
    \label{fig:3}
\end{figure}
\subsection{Parameters analysis}
For SGHormer, there are several critical hyperparameters like the membrane potential threshold $V_{th}$ and number of time steps $T$, will directly affect the performances of SGHormer. In this section, we deploy experiments on these parameters to explore the correlations between predictive performances and energy consumption of model with different parameter settings. The results are depicted in Figure \ref{fig:3}. We visualize the changes of metrics as parameters change and scale the theoretical energy consumption to a similar interval.
For the membrane potential threshold, it is directly connected with the spiking rate of whole model. As the threshold increases, the spiking rate of spiking neurons and energy consumption both decreases. We observe that optimal choices of threshold is quite different on different datasets. This is one of reasons why some parameterized LIF, which can automatically adjust membrane-related parameters, may achieve better performances.
The number of time steps is essential to approximate the real-valued inputs in rate-based SNNs. Theoretically, the firing rate can represent the original real value accurately as the number of time steps goes to infinity. We observe that The advantage of increasing the time steps for SGHormer is more in terms of speed of convergence rather than its performance. 
It make SGHormer can achieve good performance with few time steps while maintain a low-level energy consumption. Compared with SATT, SGHormer is more energy-efficient on large-scale graph data due to the simplifying attention computation.  

\section{Conclusion}
In this study, we have explored a energy-saving graph transformer driven by SNN. In order to create the fusion of GT and SNN, we design spiking rectify block (SRB) and spiking graph self-attention (SGSA). SRB enables SGHormer to maintain representation power while let spikes as the communication signals among layers. And SGSA partly alleviates internal drawbacks of SNNs on strong dependencies of time steps during calculating the attention scores. Massive experiments conducted to on graph-level benchmarks show that well-designed spiking-based GTs can bridge the performance gaps and achieve the comparable performances with extremely low energy consumption. As a energy-saving solutions from the perspective of the biological structure, SGHormer have potential to pave the path for deploying GTs on edge devices.

\bibliographystyle{named}
\bibliography{ref}

\end{document}